\documentclass[11pt]{article}
\usepackage{graphicx}
\usepackage{mathptmx}      
\usepackage{url}

\usepackage{fp,xcolor,colortbl}
\usepackage[ruled,linesnumbered]{algorithm2e}
\usepackage{amssymb}
\usepackage{booktabs}
\usepackage{lmodern}
\usepackage{mathtools}
\usepackage{subfigure}
\usepackage{multirow}

\newcommand{\bs}{\boldsymbol}

\newcommand\C{{\bs C}}

\newcommand\f{{\bs f}}

\newcommand\h{{\bs h}}

\newcommand\X{{\bs X}}
\newcommand\Y{{\bs Y}}

\newcommand\x{{\bs x}}
\newcommand\y{{\bs y}}
\newcommand\z{{\bs z}}

\newcommand\M{{\bs M}}

\newcommand\W{{\bs W}}

\newcommand{\eqdef}{\overset{\mathrm{def}}{=\joinrel=}}

\newcommand\sgn{\operatorname{sgn}}
\newcommand\etal{{\em et al.\/}\,}

\newtheorem{definition}{Definition}
\newtheorem{conjecture}{Conjecture}
\newtheorem{proposition}{Proposition}

\newcommand{\twocell}[2][c]{%
  \begin{tabular}[#1]{@{}c@{}}#2\end{tabular}}

\DontPrintSemicolon
\begin{document}
\setcounter{page}{1}

\title{Balanced Quantization: An Effective and Efficient Approach to Quantized Neural Networks}

\author{
Shuchang Zhou$^1{}^2{}^3$ , Yuzhi Wang$^3$, He Wen$^3$ , Qinyao He$^3$  and Yuheng Zou$^3$ \\
$^1$University of Chinese Academy of Sciences,\\
Beijing 100049, China\\
$^2$State Key Laboratory of Computer Architecture,\\
Institute of Computing Technology, \\
Chinese Academy of Sciences, \\
Beijing 100190, China\\
$^3$ Megvii Inc., Beijing 100190, China \\
	\texttt{shuchang.zhou@gmail.com, yz-wang12@mails.tsinghua.edu.cn, }\\
	\texttt{\{wenhe,hqy,zouyuheng\}@megvii.com}
}

\date{December 20, 2016; revised Mar. 5, 2017}
\maketitle

\begin{abstract}
Quantized Neural Networks (QNNs), which use low bitwidth numbers for representing parameters and performing computations, have been proposed to reduce the computation complexity, storage size and memory usage. In QNNs, parameters and activations are uniformly quantized, such that the multiplications and additions can be accelerated by bitwise operations. However, distributions of parameters in Neural Networks are often imbalanced, such that the uniform quantization determined from extremal values may under utilize available bitwidth. In this paper, we propose a novel quantization method that can ensure the balance of distributions of quantized values. Our method first recursively partitions the parameters by percentiles into balanced bins, and then applies uniform quantization. We also introduce computationally cheaper approximations of percentiles to reduce the computation overhead introduced. Overall, our method improves the prediction accuracies of QNNs without introducing extra computation during inference, has negligible impact on training speed, and is applicable to both Convolutional Neural Networks and Recurrent Neural Networks. Experiments on standard datasets including ImageNet and Penn Treebank confirm the effectiveness of our method. On ImageNet, the top-5 error rate of our 4-bit quantized GoogLeNet model is 12.7\%, which is superior to the state-of-the-arts of QNNs.

\end{abstract}


\normalsize
\section{Introduction}
\label{sec:intro}
Deep Neural Networks~(DNNs) have attracted considerable research interests over the
past decade.
In various applications, including computer
vision~\cite{krizhevsky2012imagenet,zeiler2014visualizing,girshick2014rich,long2015fully},
speech recognition~\cite{hinton2012deep,DBLP:conf/icassp/GravesMH13}, natural language
processing~\cite{mikolov2013distributed,sutskever2014sequence,bahdanau2014neural},
and computer games~\cite{mnih2015human,silver2016mastering}, DNNs have demonstrated their
ability to model nonlinear relationships in massive amount of data and their robustness to realworld noise.
However, the modeling capacities of DNNs are roughly proportional to their
computational complexity and number of parameters~\cite{he2016identity}. Hence
many DNNs, like VGGNet~\cite{Simonyan14c}, GoogLeNet~\cite{szegedy2014going} and
ResNet~\cite{he2015deep}, which are widely used in computer vision applications,
require billions of multiply-accumulate~operations~(MACs) even for an input image of width and
height of 224.
Moreover, as these DNN models use many channels of activations (feature maps)
for intermediate representations, they have a large runtime memory footprint
and storage size.
Such vast amount of resource requirement impedes the adoption of DNNs on devices
with limited computation resource and power supply~\cite{DBLP:journals/tc/GalalH11}, and in
user-interactive scenarios where instant responses are expected.
A similar argument also applies to Recurrent Neural Networks~(RNNs). In particular, the transition and embedding matrices in a Long
Short Time Memory~(LSTM)~\cite{hochreiter1997long} or a
Gated Recurrent Units~(GRU)~\cite{chung2014empirical} model
have dense connections that make them particularly demanding in both computation
and storage.


Many approaches have been proposed to accelerate the computation or
reduce the memory footprint and storage size of DNNs. One approach from the
hardware perspective is designing hardware accelerators for the computationally
expensive operations in DNNs~\cite{pham2012neuflow,chen2014diannao,DBLP:journals/tc/LuoLLWZCXTC17}.
From the algorithmic perspective,
a popular route to faster and smaller models is to impose constraints on the
parameters of a DNN to reduce the
number of free parameters and computational complexity, like
low-rankness~\cite{denton2014exploiting,jaderberg2014speeding,tai2015convolutional,zhou2015exploiting,novikov2015tensorizing,zhang2015accelerating},
sparsity~\cite{anwar2015structured,han2015learning,han2015deep,liu2015sparse},
circulant property~\cite{cheng2015exploration}, and sharing of
weights~\cite{chen2015compressing,DBLP:conf/kdd/ChenWTWC16}.
However, these methods use high bitwidth numbers for computations in general,
which requires availability of high precision MAC
instructions that incur high hardware complexity~\cite{anguita2011fpga}. In contrast, several previous works
have demonstrated that low bitwidth numbers may be sufficient for performing inferences with DNNs.
For example, in~\cite{vanhoucke2011improving,alvarez2016efficient,zen2016fast}, trained
DNNs are quantized to use 8-bit numbers for storing parameters and performing
computations, without incurring significant degradation of predition quality.
Gong~\etal~\cite{gong2014compressing} also applied vector quantization to
speed up inferences of DNNs. However, these works~\cite{vanhoucke2011improving,alvarez2016efficient,zen2016fast,gong2014compressing} did not integrate the
quantization operations into the training process of a DNNs, as the discrete
quantized values necessarily would have zero gradients, which would break the
Back-Propagation~(BP) algorithm. Applying quantization as a post-processing step is far from satisfactory as the quantized DNNs do not have a chance to adapt to the
quantization errors~\cite{merolla2016deep}. Consequently, 8-bit was generally
taken to be a limit for post-training quantization of DNNs~\cite{gupta2015deep}.

Recently,
Quantized Neural Networks~(QNNs)~%
\cite{DBLP:journals/corr/CourbariauxB16,wu2015quantized,kim2016bitwise,DBLP:conf/nips/HubaraCSEB16,rastegari2016xnor}
have been proposed to further reduce the bitwidths of DNNs, by incorporating
quantization into the training process. The key enabling technique is a
trick called Straight Through
Estimator~(STE)~\cite{hinton2012neural,bengio2013estimating,hwang2014fixed}, which is
based on the following observation:
as the quantized value is an approximation of the original value, we can substitute the gradient with respect
to the quantized value for the gradient of original value. Simple as it is, the
trick allows the inclusion of quantization into the computation graph of BP and
allows QNNs to represent parameters, activations and gradients with low bitwidth
numbers. The QNN technique has been successfully applied to both CNNs and
RNNs~\cite{shin2016fixed,hubara2016quantized}, to successfully produce lower
bitwidth versions of AlexNet, ResNet-18 and GoogLeNet that have comparable prediction
accuracies as their floating point counterparts. However, the
degradation of prediction accuracy is still significant for most QNNs,
especially when quantizing to less than
4-bit~\cite{miyashita2016convolutional,zhou2016dorefa}.

In this paper, we propose a Balanced Quantization method that improves the
prediction accuracies of QNNs. In general, QNNs employ uniform quantization to
eliminate floating point operations during the inference process by exploiting
the bitwise operations. However, the parameters of neural networks often
have a bell-shaped distribution and sporadic large outliers, making the quantized
values not evenly distributed among possible values when uniform quantization
is applied. In the extreme case, some of the possible quantized values are
never used.
To remedy this, we propose to use a novel quantization method that ensures the
balanced distribution of quantized values.


This paper makes the following contributions:
\begin{enumerate}

  \item We propose a Balanced Quantization
  method for the quantization of parameters of QNNs. The method emphasizes on producing balanced distributions of quantized
  values rather than preserving extremal values, by using percentiles as
  quantization thresholds. As a result, effective bitwidths of quantized
  models are increased. (See Subsection~\ref{subsec:bal-quant})
  \item To reduce the computation overhead introduced by computing percentiles,
  we approximate medians by means, which are computationally more efficient on
  existing hardware. The efficacy of the approximation is empirically validated.
  (see Subsection~\ref{sec:approx-median})
  \item Experiments confirm that
our method significantly improves the prediction accuracies of CNNs and RNNs on
standard datasets like ImageNet and Penn Treebank. (see Section~\ref{sec:experiment})
  \item The implementation of Balanced Quantization will be
  available on-line, in TensorFlow~\cite{abaditensorflow} framework.
\end{enumerate}

\section{Quantized Neural Networks}
In this section we introduce the notations and algorithms of QNNs. We also show
how QNNs can exploit bitwise operations for speeding up computations and how
to incorporate quantization steps into computation graphs
of QNNs during training.

\label{sec:nn-quant}
\subsection{Notations}

We will use the rounding operation intensively in this paper. For
tie-breaking, we apply the ``round half towards zero'' rule, which rounds positive numbers with
fraction $\frac12$ down and negative numbers with fraction $\frac12$ up. We
assigns the name of ``round-to-zero'' for this variant of rounding:
\begin{align*}
\textit{round-to-zero}(x)\eqdef \sgn(x)\left \lceil{|x|-\frac12}\right \rceil
\text{.}
\end{align*}

Without loss of generality, we represent weight parameters of a neural
network as a matrix $\W$. When doing $k$-bit uniform quantization with the step
length $\frac1{2^k-1}$, we can define a utility function $Q_\mathrm{k}$
that converts floating point numbers in close interval $[0,\;1]$ to fixed point
numbers as follows:
.
\begin{align}
\label{eq:def-q-k}
Q_\mathrm{k}(\W) \eqdef
& \frac{\textit{round-to-zero}((2^k-1)\W)}{2^k-1},\notag\\
& \quad 0\le w_{i,j} \le 1\;\forall i,j
\text{.}
\end{align}

The outputs of $Q_\mathrm{k}$ are the fixed point values $
 0,\; \frac{1}{2^k-1},\; \frac{2}{2^k-1}, \cdots ,\; 1 \text{.}$

In a Quantized Neural Network, we use $Q_\mathrm{k}$ for the quantization of
parameters, activation and gradients.
When quantizing parameters, as the utility function $Q_\mathrm{k}$ requires
the input to be in close interval $[0,\;1]$, we
should first map the parameters $\W$ to that value range.
%
The method in \cite{hubara2016quantized,zhou2016dorefa} uses the following affine transform to
change the value range:

\begin{definition} [$k$-bit Uniform Quantization]
\label{def:qk}
\begin{align*}
\varphi(\W) &\eqdef \frac{\W}{2\max(|\W|)}+\frac12 \\
\textit{quant}_\mathrm{k}(\W) & \eqdef
\varphi^{-1}(Q_\mathrm{k}(\varphi(\W)))
\text{,}
\end{align*}
\end{definition}

{\it
where the subscript $k$ in $\textit{quant}_k$ stands for $k$-bit
quantization, and $|\W|$ is a matrix with values being the absolute values of corresponding entries in
$\W$.
}

As $-\max(|\W|)\le w_{i,j}\le\max(|\W|)$, we have
$0\le\frac{w_{i,j}}{2\max(|\W|)}+\frac12\le1$. We can then apply
$Q_\mathrm{k}$ to get the fixed point values
$Q_\mathrm{k}(\varphi(\W))$, which are affine transformed by
$\varphi^{-1}$ to restore the value range back to the closed interval $[-\max(|\W|), \max(|\W|)]$.


\subsection{Simplistic View of Quantized Neural Network}
QNNs are Neural Networks that use quantized values for computations. Because the
convolutions between inputs and and convolution kernels can also be represented
as matrix products, w.l.o.g., we take a Multi-Layer Perceptron~(MLP) as an
example through out the rest of this paper.

Let the outputs,
 activation function, weight parameters and bias parameters of the $i$-th layer of
 a neural network be $\X_{i}$, $\sigma_i$, $\W_i$ and $\bs{b}_i$,
 respectively. The $i$-th Convolution/Fully-Connected layer can be represented as:
\begin{align*}
\X_{i} = \sigma_i(\W_i \X_{i-1} + \bs{b}_i)
\text{.}
\end{align*}

The corresponding formula for the $i$-th Convolution/Fully-Connected layer of a
QNN is:
\begin{align}
\label{eq:quant-nn-step}
\X_{i}^\mathrm{q} = \textit{Q}_\mathrm{A}(\sigma_i(\W^\mathrm{q}_i \X_{i-1}^\mathrm{q} + \bs{b}_i))\notag
\\
\W^\mathrm{q}_i = \textit{Q}_\mathrm{W}(\W_i)
\text{,}
\end{align}
where $\W^\mathrm{q}_i$ and $\X_{i}^\mathrm{q}$ are quantized weights and activations
respectively; $Q_\mathrm{W}$ and $Q_\mathrm{A}$ are quantization functions. Note the
bias parameters $\bs{b}_i$ may not need be quantized, for reasons we will explain in
Appendix~\ref{subsec:free-affine}.

Other types of layers like pooling layers may also take quantized values as
inputs and outputs.
The input to the first layer of a QNN may have higher bitwidth than
the rest of the network to preserve information \cite{hubara2016quantized}.

\subsection{Exploiting Bitwise Operations in QNN}
\label{sec:exploit-bit}
Using quantized values for computation makes it possible to use fixed point
operations instead of floating point operations.
We next show how to perform dot products between quantized numbers by
bitwise operations.

We first consider the dot products between $k$-bit fixed point numbers.
In the extreme case of
$k=1$, dot products are done between bit strings, which allows for the
following method of using bitwise operations:
\begin{align*}
\bs{x} \cdot \bs{y} =
\textit{bitcount}(\textit{and}(\bs{x}, \bs{y}))
\text{, } \forall i, x_i, y_i \in \{0, 1\}\text{,}
\end{align*}
{\it
where ``bitcount'' counts the number of $1$ in a bit string, and ``and'' performs bitwise AND operation.
}

In the multi-bit case ($k>1$),
we may also exploit the above kernel as in~\cite{DBLP:journals/corr/CourbariauxB16}.
Assume $\bs{x}$ is a sequence of $M$-bit fixed point integers
s.t.
$\bs{x} = \sum_{m=0}^{M-1} c_m(\bs{x}) 2^m$ and $\bs{y}$ is a
sequence of $K$-bit fixed point integers s.t. $\bs{y} = \sum_{k=0}^{K-1}
c_k(\bs{y}) 2^k$ where $(c_m(\bs{x}))_{m=0}^{M-1}$ and
$(c_k(\bs{y}))_{k=0}^{K-1}$ are bit vectors, the dot product of $\bs{x}$
and $\bs{y}$ can be computed by bitwise operations as:
\begin{align*}
\bs{x}\cdot \bs{y} = \sum_{m=0}^{M-1} \sum_{k=0}^{K-1} 2^{m+k} \,
\textit{bitcount}[\textit{and}(c_m(\bs{x}), c_k(\bs{y}))]
\text{, } \notag\\ c_m(\bs{x})_i, c_k(\bs{y})_i \in \{0, 1\} \, \forall
i,\,m,\,k\notag
\text{.}
\end{align*}

In the above equation, the computation complexity is
$O(MK)$, i.e., directly proportional to the product of bitwidths of $\bs{x}$
and $\bs{y}$. Hence it is beneficial to reduce the bitwidth of a QNN as long
as the prediction accuracy is kept at the same level.
It has been demonstrated that exploiting dot-product kernels allows for
efficient software~\cite{hubara2016quantized} and hardware
implementations~\cite{andri2016yodann,lee2016fpga}.

In Formula~\ref{eq:quant-nn-step}, the matrix multiplication happens between the
quantized values $\W^\mathrm{q}_i$ and $\X_{i-1}^\mathrm{q}$.
When the activation function is monotone, the computation of $\X_i^\mathrm{q}$ can all be performed by operations on fixed-point numbers,
even when the bias parameters $\bs{b}_i$ are floating point numbers. The method is
detailed in Appendix~\ref{subsec:free-affine}.

\subsection{Training Quantized Neural Networks by Straight-Through Estimator}

Having quantization steps in computation prevents direct training of QNNs with
the BP algorithm, as mathematically any quantization function will
have zero derivatives.
To remedy this, Courbariaux~\etal~\cite{courbariaux2015binaryconnect}
proposed to use STE to assign non-zero
gradients for quantization functions.
As the discrete parameters cannot be used to accumulate the high precision gradients, they kept two copies of parameters, one consisting of
quantized values $\W_\mathrm{q}$ and the other consisting of real values $\W$. The real
value version $\W$ is used for accumulation, while $\W_\mathrm{q}$ is used
for computation in forward and backward passes. We will refer to $\W_\mathrm{q}$ as
quantized parameters, or simply parameters, of QNNs, and reserve $\W$ for the
``floating point copy'' in the rest of this paper.
%
%

As STE introduces approximation noises into computations of gradients, we would
like to limit it to places where necessary. It can be observed that the only
function in Formula~\ref{eq:def-q-k} that has the zero gradients is the rounding
function.
Hence we construct its STE version, $\textit{round-to-zero}_{\mathrm{ste}}$, as follows:

\begin{align*}
\text{Forward:   }& \tilde{\W} \leftarrow \textit{round-to-zero}(\W) \\
\text{Backward:   }& \frac{\partial{C}}{\partial{\W}} \leftarrow
\frac{\partial{C}}{\partial{\tilde{\W}}} \text{,}
\end{align*}
where $\tilde{\W}$ is the rounded value and $C$ is the objective function used in
training of the neural network.

Functions using $\textit{round-to-zero}$ function, like the $k$-bit
uniform quantization function $\textit{quant}_\mathrm{k}$, can be transformed
into the STE version by replacing $\textit{round-to-zero}$ with
$\textit{round-to-zero}_{\mathrm{ste}}$.

A QNN can then use ${quant}_{\mathrm{ste}}$ to include quantization in its
computation graph. For completeness, we provide the inference and training algorithm of an $L$-layer QNN as Algorithm~\ref{alg:train-qnn} in Appendix~\ref{subsec:qnn-train-algo}.

\section{Balanced Quantization for Neural Network Parameters}
\label{sec:main}

In this section, we focus on more effective quantization of parameters of QNNs
to improve their prediction accuracies. We propose the Balanced
Quantization method, which induces the quantized parameters to have balanced
distributions. The method divides parameters by percentiles into bins containing the same number of
entries before the quantization.
We also propose to use approximate thresholds in the algorithm to reduce
computation overhead during training.

\subsection{Effective Bitwidth and Prediction Accuracy of QNN}
\label{subsec:error}
Using QNNs can reduce computation resource requirements considerably. However,
QNNs usually have lower prediction accuracies than their floating point
counterparts, especially when bitwidths goes below
$4$-bit~\cite{hubara2016quantized,miyashita2016convolutional,zhou2016dorefa}.

\begin{figure}[htpb]
  \centering
    \includegraphics[width=.7\linewidth]{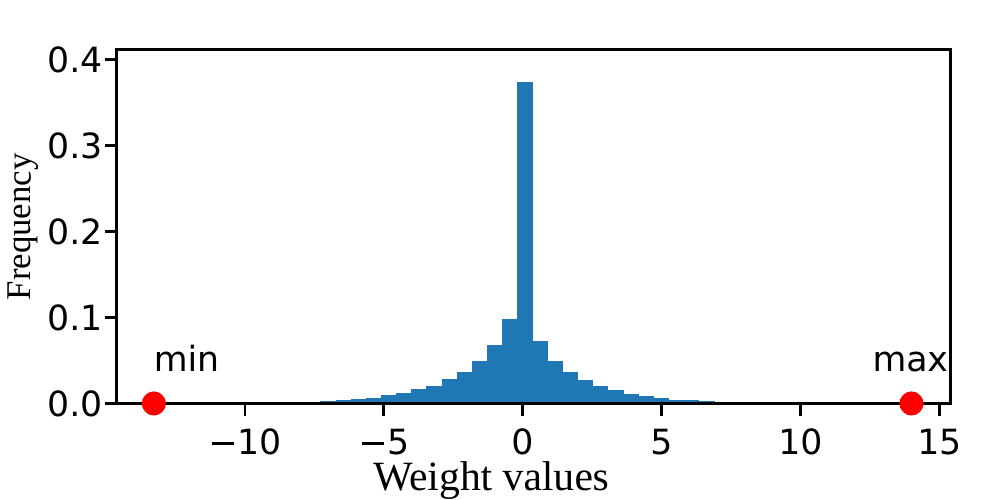}
  \caption{Floating point copy of weights in a QNN after 60 epochs of training. The weight values follow a bell-shaped distribution, and the minimum and maximum values differ a lot from the other values.}
   \label{fig:bell-shape}
\end{figure}

\begin{figure}[htpb]
	\centering
    \includegraphics[width=.7\linewidth]{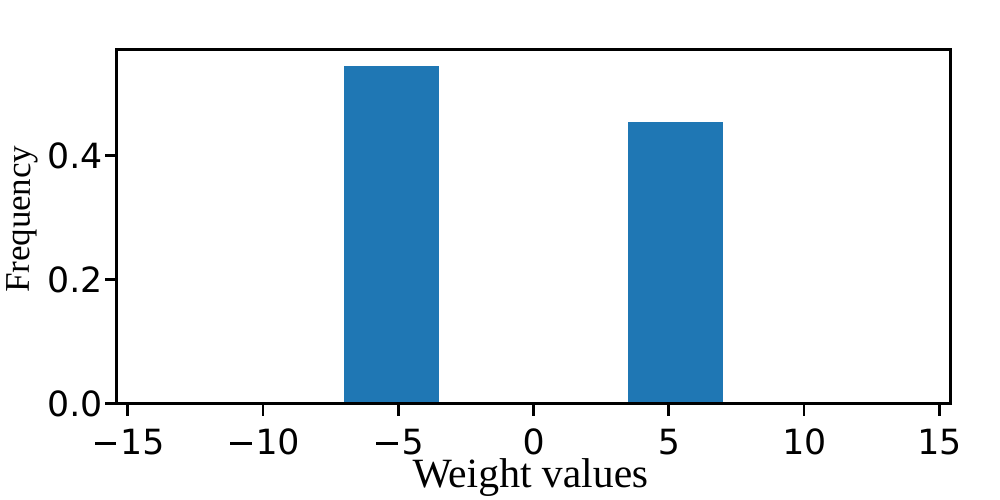}
\caption{Results of imbalanced quantization (no equalization). After uniform quantization of weight values to 2-bit numbers, the quantized values concentrate on the central two out of four possible quantized values.}
   \label{fig:bell-shape-imbal}  	
\end{figure}

We investigate this inefficiency of using low bitwidth parameters by
inspecting the parameters of QNNs before and after the quantization. On many such
models, we observe that
the parameters before the quantization follow bell-shaped distributions, just as other
DNNs~\cite{DBLP:conf/icml/SaxeKCBSN11,giryes2015deep}. Moreover, it is not rare
to observe outliers. Consequently, the quantized
values after uniform quantization will often follow imbalanced distributions between
possible values. An illustrative example is given in
Fig.~\ref{fig:bell-shape} and Fig.~\ref{fig:bell-shape-imbal},
where histograms of weight parameters, before and after the quantization, of a
layer in a quantized ResNet model are shown. The quantized weights are 2-bit.

A QNN with parameters following imbalanced distributions may be suboptimal.
For example, the 2-bit weight model in Fig.~\ref{fig:bell-shape-imbal} fails to
exploit available value range, and may be well
approximated with a 1-bit weight model.

Hence the ``real'' bitwidth of a QNN may be well below its specified bitwidth.
To quantitatively measure the ``effective'' bitwidth, we propose to use
the mean of entropy of parameters of each layer in a QNN as an indicator as
follows.

\begin{definition}
\label{def:effective-bitwidth}
\begin{align}
\textit{effective-bitwidth}(\x) \eqdef \textit{entropy}({\mathbf
P}(\x))\notag\\
=
 \textit{bitwidth} \times
\frac{\textit{entropy}({\mathbf
P}(\x))}{\textit{entropy}(\textit{UniformDistribution})}\notag
\text{,}
\end{align}
\end{definition}
{\it
where $\textit{entropy}$ is defined with base-$2$ logarithm, and ${\mathbf
P}(\x)$ refers to the distribution of $\x$.
}

The definition is in agreement with the following intuitions.
\begin{enumerate}
  \item If the quantized values are concentrated in a few bins like in
  Fig.~\ref{fig:bell-shape-imbal}, indicating poor utilization of the available
  bitwdith, the Effective Bitwidth will be low, just as expected.
  \item When the order of the bars standing for quantized values
  in histogram is permuted, which does not increase bitwidth utilization, the
  Effective Bitwidth will not change.
  \item If $\x$ is drawn from a discrete uniform distribution with
  $2^{B}$ possible values, then
  $\textit{effective-bitwidth}(\x) = {B}$ as desired.
\end{enumerate}

Based on this definition of Effective Bitwidth, we make the following
conjecture that will be empirically validated in
Subsection~\ref{subsec:entropy-accuracy}:
\begin{conjecture}
\label{conj:model-capability}
Assume other factors affecting prediction accuracies, like
learning rate schedules and model architectures, are kept the same.
The prediction accuracy of a converged QNN model is positively correlated with its
Effective Bitwidth.
\end{conjecture}

Motivated by the conjecture, we propose
a novel quantization algorithm that can enforce the balanced distribution of
quantized parameters, which maximizes entropy of the converged model and
consequently its Effective Bitwidth.

\subsection{Balanced Quantization Algorithm}
\label{subsec:bal-quant}

\subsubsection{Outline}

In this subsection we propose an algorithm to induce parameters of QNNs to have
more balanced distributions, and consequently larger Effective Bitwidths.

The first step is histogram equalization,
which can be implemented as a piecewise linear transform. The second step performs quantization, and then matches the value
range with that of the input by an affine transformation.

\begin{figure*}[!ht]
	\centering
	\includegraphics[width=\linewidth]{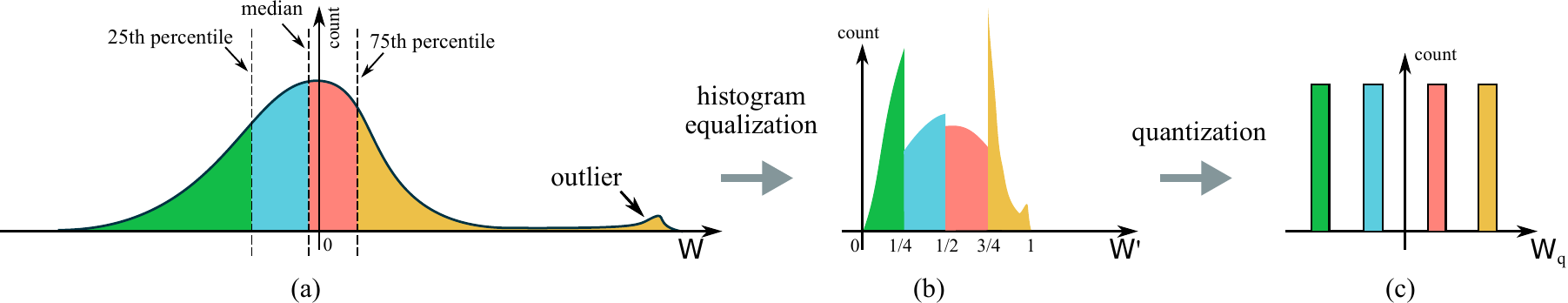}
	\caption{The schematic description of the Balanced Quantization algorithm in presence of outliers, with the case of $k=2$ as an example. The histogram of the weight values is first equalized by piecewise linear transform and then mapped to a symmetric distribution. The subfigures are (a) the histogram of floating-point weight values, (b) the histogram-equalized weight values, and (c) the quantized weight values.}
	\label{fig:bal-quant}
\end{figure*}

Fig.~\ref{fig:bal-quant} gives a schematic diagram of the Balanced
Quantization method.  The method starts by partitioning numbers into bins
containing the same number of entries. Each
partition is then mapped to a evenly-divided interval in the closed interval $[0,\;1]$. Finally, the quantization step maps intervals into discrete values
and transforms the value range to be approximately the same as input. There will
be exactly the same number of quantized values assigned to possible choices
when percentiles are used as thresholds.

Algorithm~\ref{alg:bal-quant} gives a more rigorous description of the whole process.

\begin{algorithm*}[tbp]
  \SetKwComment{comment}{\{}{\}}
  \SetNoFillComment
  \SetKwInOut{REQUIRE}{Require}
  \SetKwInOut{ENSURE}{Ensure}

  \caption{$k$-bit Balanced Quantization Algorithm of Matrix $\W$}
  \label{alg:bal-quant}
  \REQUIRE{$\W$ is a real matrix}
  \ENSURE{$\W_\mathrm{q}$ is quantized weights.}
  \BlankLine
  $\textit{scale} \leftarrow \max(|\W|)$\;
  \BlankLine
  \comment{Histogram Equalization}
  \comment{The equalized values $\W_\mathrm{e}$ are in closed interval $[0,\;1]$.}
 $\W_\mathrm{e} \leftarrow \textit{equalize}_\mathrm{k}(\W)$\;


  \BlankLine
  \comment{Quantization and restoring value range}
  \comment{$\W_\mathrm{f}$ are fixed point numbers among $2^k$ discrete values
  $-\frac12,\,-\frac12+\frac{1}{2^k-1},\,-\frac12+\frac{2}{2^k-1}\,\cdots,\,\frac12$.}

  $\W_\mathrm{f} \leftarrow
  \frac{1}{2^k-1}\textit{round-to-zero}(2^{k}\W_\mathrm{e} - \frac12)-
  \frac{1}{2}$\;

  \comment{Values of $\W_\mathrm{q}$ are scaled fixed point numbers in
  closed interval $[-\max(|\W|),\;\max(|\W|)]$.}
  $\W_\mathrm{q} \leftarrow 2\times\textit{scale}\times\W_\mathrm{f}$\;

  \BlankLine

\end{algorithm*}

\subsubsection{Histogram Equalization by Piecewise Linear Transform}
In this subsection,
we detail the histogram equalization step, which we adapt from image
processing literature~\cite{DBLP:conf/siggraph/Heckbert82}.

Assume we are quantizing to $k$-bits values and $N=2^k$. The input value range is divided to $N$ intervals, including $N-1$ number of
half open intervals $[t_i, t_{i+1})$ and a closed interval $[t_{N-1}, t_N]$. To simplify notation, we denote the
$i$-th interval as $I_i$. Thresholds $\{t_i\}_{i=0}^N$ are determined by the
algorithm of histogram equalization. When exact equalization is desired, we let thresholds
$t_i$ be the $\frac{100 i}{N}$-th percentiles of the original
distribution.
The formula of equalized values $x_e$ is as follows.
\begin{definition}[Histogram equalization]
\label{def:hist-eq}
\begin{align*}
x_\mathrm{e} = {equalize}_{\mathrm{k}}(x) \eqdef  a_i x + b_i \notag\\
\text{ if }
x \in I_i,\quad 0\le i\le N-1,\,i\in \mathbb{Z}.
\end{align*}
\end{definition}

As $equalize_\mathrm{k}$ maps $I_i$ to evenly spaced segments $J_i$ of
target interval $[0, 1]$, parameters of the affine transformations, $a_i$ and $b_i$, can be determined
from the following constraints:
\begin{align*}
a_i t_i + b_i & = \frac{i}{N},\\
a_i t_{i+1} + b_i & = \frac{i+1}{N},\quad 0\le i\le
N-1,
\end{align*}
where $\frac{i}{N}$ and $\frac{i+1}{N}$ are the two endpoints of $J_i$.

Let $C$ be the objective function of the training, the back-propagation formula
for $equalize_\mathrm{k}$ is straightforward:
\begin{align*}
& \frac{1}{a_i}\frac{\partial{C}}{\partial x}  = \frac{\partial{C}}{\partial x_\mathrm{e}} \notag\\
& \text{ if } x_\mathrm{e} \in J_i,\quad 0\le i\le N-1,\,i\in \mathbb{Z}.
\end{align*}

\begin{algorithm*}[tbp]
  \SetKwComment{comment}{\{}{\}}
  \SetNoFillComment
  \SetKw{Fn}{Function}{}
  \SetKwFunction{uQuantize}{HistogramEqualize}

  \caption{Histogram Equalization of Matrix $\W$ by Recursive Partitioning}
  \label{alg:recursive-qnn}

  \BlankLine
  \Fn{\uQuantize{$\W$, $\M$, $\textit{level}$}}{
  \KwData{\\
    \quad $\W$ is a real-valued matrix;\\
    \quad $\M$ is a mask matrix with values in $\{0, 1\}$ and has the same shape
    as $\W$; It is used to note the ``working set'' of $\W$.\\
    \quad $\textit{level}$ is an auxiliary variable recording recursion
    level.
  }
  \KwResult{A matrix of the same shape as $\W$ with value range $[0,\;1]$}

  \BlankLine

  \comment{$S_\W$ is the subset of the elements of $\W$ with positive masks.}
  $S_\W \gets \{w_{i,j} | \,\forall\,\, w_{i,j} \in \W; m_{i,j} > 0\}$ \;

  \If{$\textit{level} = 0$}{
    \comment{Affine transform $\W$ to the value range of $[0, 1]$.}
    \comment{$*$ is element-wise (Hadamard) multiplication.}
    \KwRet $\dfrac{\W-\min(S_\W)}{\max(S_\W)-\min(S_\W)} \circ \M$ \;
  }
  \BlankLine
  \comment{Construct two masks $\M^l$ and $\M^g$ using $\mathrm{mean}(S_\W)$ as threshold.}
  \comment{$\mathrm{mean}(S_\W)$ is used to replace $\mathrm{median}(S_\W)$ so
  as to accelerate computation (see Section~\ref{sec:approx-median}).}
  $T \gets \mathrm{mean}(S_\W) = \frac{\sum(\W \circ
\M)}{\sum\M}$ \;
  \BlankLine
  $\M^l \gets {\bs 0}$, $\M^g \gets {\bs 0}$ \;
  \For{$w_{i,j} \in \W ;\, m_{i,j} > 0$} {
    \eIf{$w_{i,j} < T$} {
      $m^l_{i,j} \gets 1$
    }{
      $m^g_{i,j} \gets 1$
    }
  }
  \label{line:mask}

  $\W^{l} \gets$ \uQuantize{$\W$, $\M^l$, $\textit{level}-1$} \;
  $\W^{g} \gets$ \uQuantize{$\W$, $\M^g$, $\textit{level}-1$} \;

  \comment{Value ranges of both $\W^{l}$ and $\W^{g}$ are $[0,\,1]$.}
  \comment{$\frac{1}{2}$ is added to $\frac{1}{2}\W^{g}$ to shift the value
  range to $[\frac12, 1]$.}
  \KwRet $\frac{1}{2}\W^{l} + (\frac{1}{2}\W^{g} +
  \frac{1}{2}) \circ \M^g$ }
  \label{line:recursive-qnn-return}
\end{algorithm*}

\subsubsection{Rounding and Restoring Value Range}
\label{subsec:round-and-restore-value}
After the histogram equalization step, the values $\W_\mathrm{e}$
are still floating point values, and need be converted to discrete values. The conversion can be done by the construction of fixed point version
$\W_\mathrm{f} = \frac{1}{2^k-1}\textit{round-to-zero}(2^k \W_\mathrm{e} - \frac12)-\frac12$. Note the mapping between $W_\mathrm{e}$ and $W_\mathrm{f}$ is different from $Q_\mathrm{k}$. For example, it maps the interval of $[0,\,\frac1{2^k}]$ to $0$,
while $Q_\mathrm{k}$ maps $[0,\,\frac1{2(2^k-1)}]$ to $0$.

Finally, $\W_\mathrm{f}$, which has value range $[-\frac12,\,\frac12]$, can be scaled
by $2\max(|\W|)$ to match the original value range.


\subsection{Approximation of Median and Efficient Implementation}
\label{sec:approx-median}
The histogram equalization defined by the piecewise linear transform in
Definition~\ref{def:hist-eq} has well-defined gradients and can be readily
integrated into the training process of QNNs. However, a naive implementation using percentiles as thresholds would
require sorting of weight values during each forward operation in BP,
which may slow down the training process of QNNs as sorting is less efficient on
modern hardware than matrix multiplications.
In this subsection, we discuss an approximate
equalization that allows efficient implementation. We first propose a recursive
implementation of histogram equalization that only requires computing medians. Noting that medians can be
well approximated by means, we construct Algorithm~\ref{alg:recursive-qnn} that
can perform approximate histogram equalization without doing sorting.


\subsubsection{Recursive Partitioning}
We first note that the $2^k$ evenly spaced percentiles required in histogram
equalization can be computed from the recursive application of partitioning
of numbers by medians. For example, when doing histogram equalization for the 2-bit
quantization, we need to compute the $25$-th, the $50$-th and the $75$-th percentiles as
thresholds. However, the $50$-th percentile is exactly the median, while the $25$-th percentile
(25\% of values are below this number) is the median of those
values that are below the median of the original distribution. Hence we can replace the computation of percentiles
with recursive applications of partitioning by medians.

\begin{figure}[htpb]
	\centering
    \includegraphics[width=.7\linewidth]{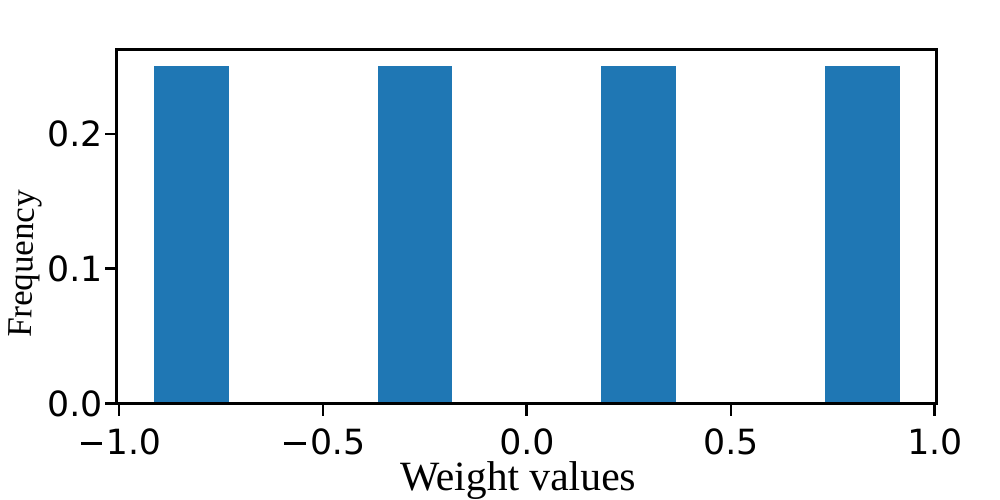}
	\caption{Balanced quantization with median (before matching value range)}
	\label{fig:median_partition}
\end{figure}

\begin{figure}[htpb]
	\centering
    \includegraphics[width=.7\linewidth]{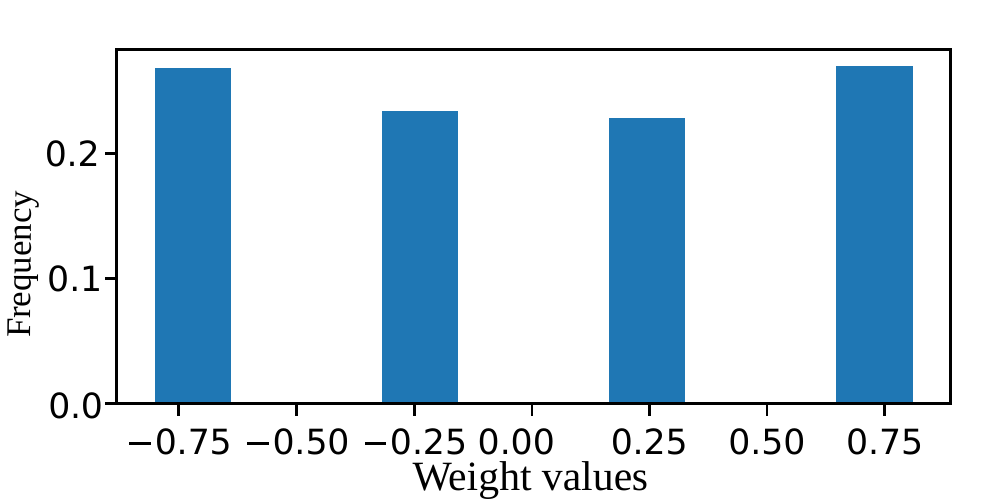}
	\caption{Balanced quantization with mean (before matching value range)}
	\label{fig:mean_partition}	
\end{figure}

Moreover, we note that when a distribution has bounded variance $\sigma$, the
mean $\mu$ approximates the median $m$ as there is an inequality bounding the difference~\cite{mallows1991another}:
\begin{align*}
|\mu - m| \le \sigma.
\end{align*}
Hence we may use means instead of medians in the recursive partitioning. The results of partitioning by different methods are shown in
Fig.~\ref{fig:median_partition} and Fig.~\ref{fig:mean_partition}. It can be
observed that partitioning by the median achieves perfect balance, and partitioning
by the mean achieves approximate balance.

\subsubsection{Implementation}
Based on the fact that the median approximates the mean, histogram equalization can be
implemented as in
Algorithm~\ref{alg:recursive-qnn}.
An auxiliary mask matrix $\M$, whose values are either $0$ or $1$, is
introduced to help manipulate the branching and selection operations. Note the
mask $\M$, which is an argument of $\operatorname{HistogramEqualize}$ at the top of call chain, is initialized
 to be ${\bs 1}$, a matrix with all values being $1$.

When Algorithm~\ref{alg:recursive-qnn} is used as the histogram equalization
step in Algorithm~\ref{alg:bal-quant}, we can prove the following proposition
(see Appendix~\ref{subsec:proof} for proof):

\begin{proposition}\label{thm:balanced-quantization-k}
\normalfont
If during application of
Algorithm~\ref{alg:recursive-qnn} the following
holds after Line~\ref{line:mask}:
\begin{align*}
\frac1{\gamma}\le \frac{\sum\M^l}{\sum\M^g} \le\gamma
\text{,}
\end{align*}
then the most frequent entry of the quantized values will appear at most
$\gamma^{2K}$ as often as that of the least frequent entry, when quantizing to
$K$-bit numbers with Algorithm~\ref{alg:bal-quant}.
\end{proposition}

\section{Experiments}
\label{sec:experiment}
In this section we empirically validate the effectiveness of the Balanced Quantization through experiments
on quantized Convolutional Neural Networks and Recurrent Neural Networks.

In our implementations of QNNs,
we convert parameters and input activations of all layers in the network to low bitwidth number, which is in line with the practice of
Hubara~\etal~\cite{hubara2016quantized}. The
CNN models used in this section are all equipped with Batch Normalization~%
\cite{ioffe2015batch} to speed up convergence. Experiments are done on Linux
machines with Intel Xeon CPUs and NVidia TitanX Graphic Processing Units.

\subsection{Experiments on Convolutional Neural Networks}
\subsection{Datasets}
For evaluation on CNNs, we conduct experiments on two datasets used for the
image classification task.

The SVHN dataset \cite{netzer2011reading}
is a real-world digit recognition dataset consisting of photos of house numbers in Google Street View images.
We consider the ``cropped'' format of the dataset: 32-by-32 colored images centered
around a single character. We also include the ``extra'' part of labeled data in training.

The ImageNet dataset contains 1.2M
images for training and 50K images for validation.
Each image in the dataset is assigned a label in one of the 1000 categories.
While testing, images are first resized such that the shortest edge is 256 pixels,
and then the center 224-by-224 crops are fed into models.
Following the conventions, we report results in two measures: single-crop top-1
error rate and top-5 error rate over ILSVRC12 validation sets~\cite{russakovsky2015imagenet}.  For brevity, we will denote the top-1 and top-5 error rates as ``top-1'' and ``top-5'', respectively.


\subsubsection{Effective Bitwdiths and Prediction Accuracies of Converged Models}
\label{subsec:entropy-accuracy}

\begin{figure}[htpb]
  \centering
      \includegraphics[width=0.7\linewidth]{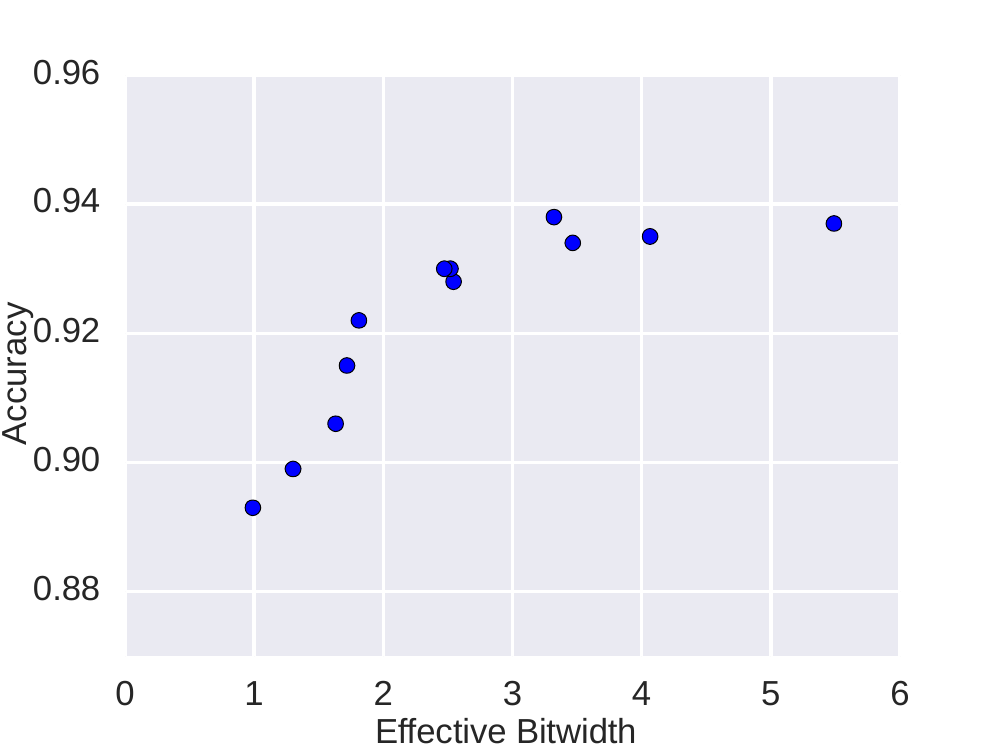}
   \caption{Relationship between Effective Bitwidths and prediction accuracies of several converged QNNs on the SVHN dataset. The models are produced by different specified bitwidths (ranging from 1-bit to 8-bit) and quantization methods (balanced or not), but all have the same architecture and training settings.}
  \label{fig:bitwidth-accuracy}
\end{figure}

\begin{figure}[htpb]
	\centering
      \includegraphics[width=0.7\linewidth]{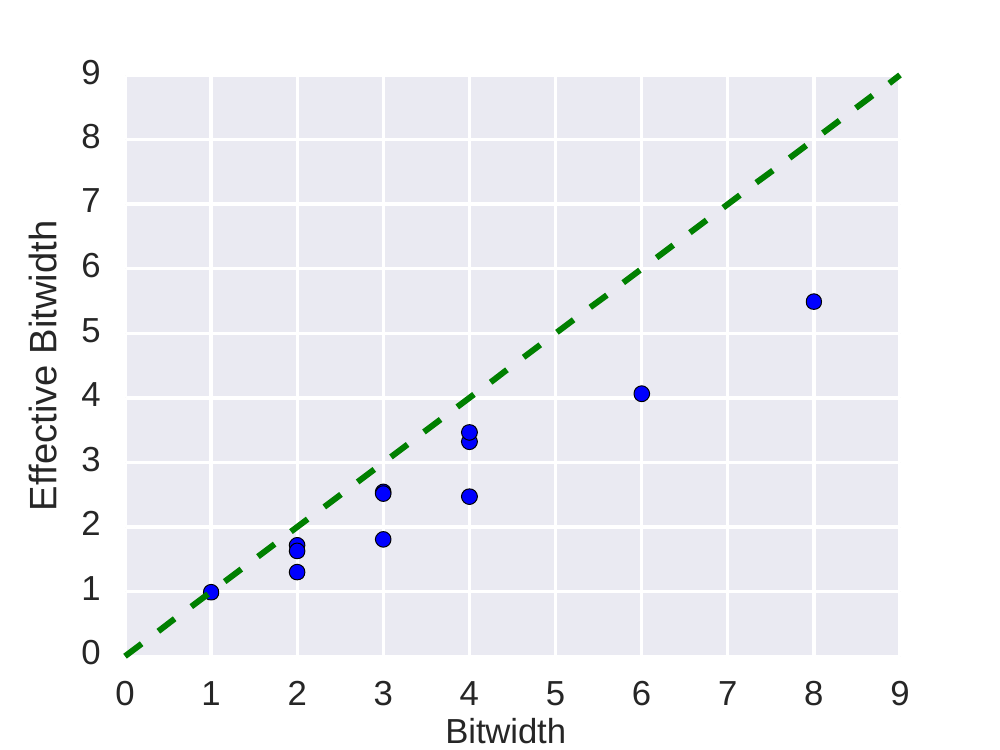}
  \caption{Relationship between Effective Bitwidth and specified Bitwidth. In general, Effective Bitwidths grow with Bitwidths. But Effective Bitwidths of most of the models are significantly less than its specified Bitwidth.}
  \label{fig:bitwidth-accuracy-alloc}
\end{figure}

In Fig.~\ref{fig:bitwidth-accuracy}
and Fig.~\ref{fig:bitwidth-accuracy-alloc}, we plot the prediction accuracies
of several converged QNNs against their Effective Bitwidths as defined in Definition~\ref{def:effective-bitwidth}.
The QNNs are trained on the SVHN dataset and have the same 7-layer CNN model
architecture; hyper-parameters like learning rate schedule, numbers of epochs are kept the same,
such that the differences between these models are only the specified bitwidths of
parameters and the quantization methods. In this way, we can evaluate the impact of Effective Bitwidths on the
prediction accuracies of converged models.

It can be observed from Fig.\ref{fig:bitwidth-accuracy} that in general,
accuracy grows with the increase of Effective Bitwidth. However, the
growth of the accuracy gradually slows down to the right half of the diagram, when the
prediction accuracy of a quantized model approaches the upper bound set
by floating point models.

\begin{table}[!ht] \centering
\caption{Evaluation of using means instead of medians when performing
Balanced Quantization, on GoogLeNet with 4-bit weights and 4-bit activations.}
\begin{center}
\begin{tabular}{cccc}
\toprule Thresholds & Top-1 & Top-5 & Effective Bitwidth\\
\midrule
mean	 & 32.3\% & 12.7\% & 3.99\\
median	 & 33.8\% & 13.3\% & 4.00 \\
\bottomrule
\end{tabular}
\end{center}
\label{tab:scaled-mean}
\end{table}

\begin{table*}[!ht]
\caption{Comparison of performances of quantized
AlexNet and ResNet. }
\begin{center}
	\begin{tabular}{p{4cm}p{1cm}p{1cm}p{1.5cm}p{1cm}p{1cm}p{1.5cm}}
\toprule
\multirow{2}{*}{\textbf{Method}} &
        \multicolumn{3}{c}{AlexNet} & \multicolumn{3}{c}{ResNet-18} \\
      \cline{2-4} \cline{5-7}
      & Top-1 & Top-5 & Effective Bitwidth & Top-1 & Top-5 & Effective Bitwidth \\
\hline
\hline
FP& 42.9\% & 20.6\% & - & 31.8\% & 12.5\% & -\\
equalized FP weights & 42.7\% & 20.9\% & - & 36.2\% & 15.3\% & - \\
FP weight + 2-bit feature & 43.5\% & 21.0\% & - & 38.9\% & 17.3\% & - \\
\hline
	imbalanced 2-bit (different settings)& 46.4\% & 24.7\% & 1.89 & 46.6\% & 22.1\% & 0.99 \\
	imbalanced 2-bit& 45.3\% & 22.3\% & 1.94 & 42.3\% & 19.2\% & 1.96\\
	balanced 2-bit& {\bf 44.3\%} & {\bf 22.0\%} & 1.99 & {\bf 40.6\%} & {\bf 18.0\%} & 1.99\\
\hline
\hline
\end{tabular}
\end{center}
\label{tab:alexnet-resnet}
FP stands for floating point. Results in rows prefixed with ``imbalanced''
are produced from direct applications of uniform quantization. Results in rows
marked with ``equalized FP weights'' only perform equalization of weights on FP
models. As the floating point values do not have well-defined effective bitwidths, we omit these entries by using the ``-'' symbols.
\end{table*}

\subsubsection{Evaluation of Approximation of Median}
In this subsection we validate the effectiveness of approximation of
the median by the mean, as proposed in Subsection~\ref{sec:approx-median}. As computing the median requires
doing sorting of weight parameters of a layer, experiments on DNNs
with many parameters will be very slow. Hence we perform experiments on
the GoogLeNet, which contains fewer than 7M parameters.
From Table~\ref{tab:scaled-mean}, it can be seen that replacing medians by means does not degrade prediction accuracies. In
fact, the method using means as thresholds is even slightly better than
the method using medians, both in terms of top-1 and top-5 error rates.

As replacing medians with means is empirically found to be viable, we will use means as thresholds in
experiments in the rest of Section~\ref{sec:experiment}.

\subsubsection{Balanced Quantization of AlexNet, ResNet-18 and GoogLeNet}
\label{subsec:imagenet}
The experiment results on AlexNet and ResNet-18 are summarized in
Table~\ref{tab:alexnet-resnet}. The results marked with ``different settings'' come from
models trained with a different learning rate schedule and clipping of weights.
It can be seen that results of the Balanced Quantization method consistently
outperform those of the uniform quantization methods (hereafter denoted as
Imbalanced Quantization) defined in Definition~\ref{def:qk}. In particular, the
top-5 error rate of the Balanced Quantized 2-bit AlexNet is within 2 percentages of that of the floating point version, making the quantized network a good candidate to replace the floating
point version in practice. As the model size
can be reduced to $\frac{1}{16}$ of the original and computations can be
performed by 2-bit numbers, the savings in resource requirements will be
significant.

However, the improvements of accuracies
due to Balanced Quantization may not be large, as accuracies of models
quantized without balance are already close to the upper
bounds set by models with floating point weights and 2-bit features.


\begin{table}[!ht] \centering
\caption{Comparison of classification error rates with state-of-the-arts on quantized
GoogLeNet model with 4-bit weights and 4-bit activations.}
\begin{center}
\begin{tabular}{ccc}
\toprule Method & Top-1 & Top-5
\\
\hline
Our float32	 & 28.5\% & 10.1\% \\
\hline
\hline
	QNN 4-bit~\cite{hubara2016quantized}   & 33.5\% & 16.6\% \\
	Ristretto 8-bit~\cite{gysel2016hardware} & 33.4\% & - \\
\hline
	Our 4-bit	  & {\bf 32.3\% } & {\bf 12.7\% } \\
\hline
\end{tabular}
\end{center}
\label{tab:googlenet-exp}
\end{table}

Table~\ref{tab:googlenet-exp} compares QNNs quantized with our method with
state-of-the-arts.
It can be seen that our method consistently
outperforms the others.
In particular, our method reduces the top-5 accuracy degradation, which is the
difference in accuracy between a QNN and a floating point version, from 6.5
percentages to 2.6 percentages.

\begin{table*}[!ht] 
\small
	\caption{Performance of Quantized RNNs on PTB datasets.}
	\begin{center}
		\begin{tabular}{
				 p{0.20\linewidth}  p{0.07\linewidth}  p{0.07\linewidth}
				p{0.10\linewidth}  p{0.11\linewidth}  p{0.10\linewidth}  p{0.11\linewidth}}
			\hline
			\multirow{2}{*}{\textbf{Model}} & \multirow{2}{*}{w-bits} & \multirow{2}{*}{a-bits} &
        \multicolumn{2}{c}{PPW} & \multicolumn{2}{c}{Effective Bitwidth} \\
      \cline{4-5} \cline{6-7}
      &  &  & balanced & imbalanced & balanced & imbalanced \\
 			\hline
      GRU & 2 & 2 & 142 & 165 & 1.98 & 1.56 \\
      GRU & 4 & 4 & 116 & 120 & 3.86 & 3.26 \\
      \twocell{GRU ($\tanh(\W)$)} & FP & FP & - & 118 & - & - \\
      GRU & FP & FP & - & 100 & - & -\\
			\hline
      LSTM & 2 & 2 & {\bf 126} & 164 & 1.96 & 1.00 \\
			\hline
      LSTM & 2 & 3 & {\bf 123} & 155 & 1.95 & 1.00 \\
      \twocell{LSTM~\cite{hubara2016quantized}} & 2 & 3 &  & 220 & & \\
			\hline			
      LSTM & 4 & 4 & 114 & 127 & 3.89 & 1.80 \\
			\twocell{LSTM~\cite{hubara2016quantized}} & 4 & 4 &  & {\bf 100} \\
			\hline					
      \twocell{LSTM ($\tanh(\W)) $} & FP & FP & - & 122 & - & -\\
      LSTM & FP & FP & - & 106 & - & - \\
			\twocell{LSTM~\cite{hubara2016quantized}} & FP & FP &  & {\bf 97} \\									
			\hline
		\end{tabular}
	\end{center}
FP stands for 32-bit floating point. Results marked with $\tanh(\W)$ are of
models that have their weights clipped by $\tanh$ before passing to
quantization. The best results for each bitwidth setting are marked in bold.
	\label{tab:bit_rnn_ptb}
\end{table*}

\subsubsection{Break-down of Accuracy Degradation with Balanced Quantization}
Overall,
the change in accuracy due to Balanced Quantization will be made up of two parts:
\begin{align}
\Delta \text{Accuracy}_{\mathrm{total}} = \Delta \text{Accuracy}_{\mathrm{eq}} + \Delta
\text{Accuracy}_{\mathrm{quant}}
\text{,}\notag
\end{align}
where $\Delta\text{Accuracy}_{\mathrm{eq}}$ stands for the change in accuracy due to
equalization and $\Delta
\text{Accuracy}_{\mathrm{quant}}$ is that of quantization.

The histogram equalization effectively imposes an additional constraint on the
neural network parameters. As the constraint limits the optimization space
of parameters, it will likely introduce additional errors into predictions of
neural networks.

Nevertheless, through the experiments in Table~\ref{tab:alexnet-resnet}, we have
observed that the reduction in $\Delta \text{Accuracy}_{\mathrm{quant}}$
outweighs the inclusion of additional term $\Delta \text{Accuracy}_{\mathrm{eq}}$.
We leave it as future work to investigate the cause and
further reduction of $\Delta \text{Accuracy}_{\mathrm{eq}}$.


\subsection{Experiments on Recurrent Neural Networks}
\label{subsec:rnn-exp}
In this subsection we evaluate the effect of Balanced Quantization on a few
Recurrent Neural Networks. We take language modeling task as an example,
and use the Penn Treebank dataset~\cite{taylor2003penn}, which contains 10K unique
words.

For fair comparison, in the following experiments, all of our models use one hidden
layer with 300 hidden units, which is the same with~\cite{hubara2016quantized}.
A word embedding layer is used at the input side of
the network whose weights are trained from scratch.
The performance is measured in perplexity per word (PPW) metric.

During experiments we find the magnitudes of weights often grow rapidly
with training when using small bitwidth, and may result in divergence.
This can be alleviated by adding $\tanh$ to constrain the value ranges
\cite{zhou2016dorefa} and adding weight decays for regularization. However, we
find using $\tanh$ to clip parameters will degrade prediction accuracy of floating point Neural
Network. Further investigation of this drop of accuracy is out of
the scope of this paper, and will be left as future work.

Experiment results are reported in Table~\ref{tab:bit_rnn_ptb}. Our result is in
agreement with~\cite{hubara2016quantized} in finding that using 4-bit
weights and activations can achieve comparable performance as floating point
counterparts.
However, we report higher accuracy than~\cite{hubara2016quantized} when using
less bits, such as 2-bit weight and activations. In particular, our 2-bit weights and 3-bit
activations LSTM achieve 155 PPW for imbalanced quantization and 123
PPW for balanced quantization, both of which outperform the counterparts
in~\cite{hubara2016quantized} by large margins, despite that our floating point models are worse than those of \cite{hubara2016quantized}.

\section{Conclusions}
\label{sec:conclusion}
In this paper, we have introduced the method of Balanced Quantization,
which enforces the quantized values to have balanced distributions through the
use of histogram equalization. Our method breaks away from traditional
quantization methods in that it emphasizes on shaping distributions of quantized
values. When incorporated into the training process of Quantized Neural Networks, our method
can improve the prediction accuracies of converged models. We have also introduced Effective
Bitwidth, which measures the utilization of bitwidth in QNNs, that can help
identify models that can benefit more from the Balanced Quantization method.

To reduce the computation overhead introduced by the need to compute percentiles
when performing Balanced Quantization, we also propose to use recursive
application of mean as approximations of percentiles (see
Subsection~\ref{sec:approx-median}).
We have also applied the
Balanced Quantization method to several popular Neural Network architectures
like AlexNet, GoogLeNet and ResNet, and found that our method outperforms the
state-of-the-arts of QNN, in terms of prediction accuracy (see
Subsection~\ref{subsec:imagenet}). Experiments on LSTM and GRU are also
encouraging (see Subsection~\ref{subsec:rnn-exp}).

As future work, it would be interesting to use the histogram transformation
technique to induce distributions that have other benefits, like a high
ratio of zeros in quantized values. It would also be interesting to
investigate whether inducing activations of neural networks to have balanced
distributions could improve the prediction accuracies of
QNNs.


\appendix
\section*{Appendix}
\label{sec:appendix}
\section{Eliminating All Floating Point Operations During Inference}
\label{subsec:free-affine}

Recall that the $i$-th layer of a QNN is like:
\begin{align*}
\X_{i}^\mathrm{q} & = Q_\mathrm{A}(\sigma_i(\W_i^\mathrm{q} \X_{i-1}^\mathrm{q} + \bs{b}_i)) \\
\W_i^\mathrm{q} & =Q_\mathrm{W}(\W_i)
\text{,}
\end{align*}
where $\sigma_i$ is the activation function, and $Q_\mathrm{A}$ and $Q_\mathrm{W}$ are
quantization functions.

Below we assume the following conditions:
\begin{enumerate}
  \item $\W_i^\mathrm{q}$ can be represented as fixed
point numbers scaled by a floating point scalar $\alpha$, i.e.\ $\W_i^\mathrm{q} = \alpha \W_\mathrm{f}$, where $\W_\mathrm{f}$ is fixed point numbers.
  \item $\X_{i-1}^\mathrm{q}$
contains only fixed point numbers
\item $\sigma_i$ is a monotone
function
\end{enumerate}
We next show that under these assumptions, the computation of $\X_{i}^\mathrm{q}$
can be done by operations between fixed point numbers.

First, by replacing the variables we have:
\begin{align*}
\X_{i}^\mathrm{q} = Q_\mathrm{A}(\sigma_i(\alpha \W_\mathrm{f} \X_{i-1}^\mathrm{q} + \bs{b}_i))
\text{.}
\end{align*}

As $Q_\mathrm{A}$ is a uniform quantization function, it can be computed by comparing
values of $\sigma_i(\alpha \W_\mathrm{f} \X_{i-1}^\mathrm{q} + \bs{b}_i)$ with a sequence of
thresholds $h_1,\,h_2,\,\cdots,h_n$. As $\sigma_i$ is monotone, and w.l.o.g.\
assume $\alpha > 0$, the comparison can equivalently be done between $\W_\mathrm{f}
\X_{i-1}^\mathrm{q}$ and
\begin{align*}
\frac1{\alpha}(\sigma_i^{-1}(h_1) - b),\,\frac1{\alpha}(\sigma_i^{-1}(h_2) -
b),\,\cdots,\frac1{\alpha}(\sigma_i^{-1}(h_n) - b)
\text{.}
\end{align*}

As $\W_\mathrm{f}$ and $\X_{i-1}^\mathrm{q}$ are fixed point numbers, their product is
necessarily made of fixed point numbers, hence there exists a sufficiently large integer
$K$ such that $2^K \W_\mathrm{f} \X_{i-1}^\mathrm{q}$ are integers. The comparison required for
computing $Q_\mathrm{A}$ can be done by comparing $2^K \W_\mathrm{f} \X_{i-1}^\mathrm{q}$ with integers
\begin{align*}
&\lfloor\frac{2^K}{\alpha}(\sigma_i^{-1}(h_1) -
b)\rfloor,\,\lfloor\frac{2^K}{\alpha}(\sigma_i^{-1}(h_2) -
b)\rfloor,\,\cdots,\notag\\
&\lfloor\frac{2^K}{\alpha}(\sigma_i^{-1}(h_n) - b)\rfloor
\text{.}
\end{align*}

Hence the computation of $\X_i^\mathrm{q}$ can be done by the comparison between fixed point
numbers $\W_\mathrm{f} \X_{i-1}^\mathrm{q}$ with the following thresholds that can be
pre-computed and stored (hence eliminating need for floating point operations
during inference):
\begin{align*}
&2^{-K}\lfloor\frac{2^K}{\alpha}(\sigma_i^{-1}(h_1) -
b)\rfloor,\,2^{-K}\lfloor\frac{2^K}{\alpha}(\sigma_i^{-1}(h_2) -
b)\rfloor,\notag\\
&\cdots,2^{-K}\lfloor\frac{2^K}{\alpha}(\sigma_i^{-1}(h_n) - b)\rfloor
\end{align*}

\section{Training Algorithm of QNN}
\label{subsec:qnn-train-algo}
For completeness we outline the training algorithm of QNNs in
Algorithm~\ref{alg:train-qnn}. Weights, activations and gradients are quantized by quantization functions
$Q_\mathrm{W}$, $Q_\mathrm{A}$ and $Q_\mathrm{G}$, that are applied to weights, activations and
gradients respectively.
$C$ stands for the cost function of the neural network. $backward\_input$ and
$backward\_weight$ are functions derived from chain rule for computing
gradients with respect to inputs and weights, respectively.
The Update function is determined by the learning rule used.
The algorithm extends Algorithm 1 in
Hubara~\etal~\cite{hubara2016quantized} to include the quantization of gradients,
and the multi-bit quantization.

\begin{algorithm*}[tbp]
  \SetKwInOut{REQUIRE}{Require}
  \SetKwInOut{ENSURE}{Ensure}
  \SetKwComment{comment}{\{}{\}}
  \SetNoFillComment

  \caption{Training a $L$-layer CNN with $W$-bit weights and $A$-bit activations
  using $G$-bit gradients.
  }
  \label{alg:train-qnn}
  \BlankLine

  \REQUIRE{a minibatch of inputs and labels $(\X_0, \Y)$, previous weights $\W$,
  learning rate $\eta$}
  \ENSURE{updated weights $\W^{t+1}$}

  \BlankLine

  \comment{1. Computing the parameter gradients:}
  \comment{1.1 Forward Propagation:}
  \For{$i = 1\, \to\, L $}{
    $\W_i^\mathrm{q} \gets Q_\mathrm{W}(\W_i)$ \;
    $\tilde{\X}_i \gets \X_{i-1}^\mathrm{q}\W_i^\mathrm{q} + \bs{b}_i$\;
    $\X_i \gets \sigma(\tilde{\X}_i)$ \;
    \If{$k < L$}{
      $\X_i^\mathrm{q} \gets Q_\mathrm{A}(\X_i)$ \;
    }
    Optionally apply pooling \;
  }

  \BlankLine

  \comment{1.2 Backward propagation:}
  Compute $g_\mathrm{L} = \frac{\partial C}{\partial \X_\mathrm{L}}$ knowing $\X_\mathrm{L}$ and label $\Y$. \;
  \For{$i = L \, \to \, 1$} {
    Back-propagate $ g_{i}$ through activation function $\sigma$ \;
    $g^\mathrm{q}_{i} \gets Q_\mathrm{G}(g_{i})$ \;
    $g_{i-1} \gets \texttt{backward\_input}(g^\mathrm{q}_{i}, \W_i^\mathrm{q})$ \;
    $g_{\W_i} \gets \texttt{backward\_weight}(g^\mathrm{q}_{i}, \X_{i-1}^\mathrm{q})$ \;
    Back-propagate gradients through pooling layer if there is one \;
  }

  \BlankLine

  \comment{2. Accumulating the parameters gradients:}
  \For{$k = 1\, \to\, L $} {
    $ g_{i} = g_{i}^\mathrm{q} \frac{\partial \W_i^\mathrm{q}}{\partial \W_i}$\;
    $\W_i^{t+1} \gets Update(\W_i, g_{i}, \eta)$\;
  }
\end{algorithm*}

\section{Proof of Proposition~\ref{thm:balanced-quantization-k}}
\label{subsec:proof}

\paragraph{Proof}
The step after histogram equalization in Algorithm~\ref{alg:bal-quant} maps the
following half open (close) intervals into quantization values:
\begin{align*}
[0,\,\frac{1}{2^K}),\,[\frac{1}{2^K},\,\frac{2}{2^K}),\,\cdots,\,[\frac{2^K-1}{2^K},\,1]
.
\end{align*}

Hence it is sufficient to prove the counting statements for these intervals
after application of Algorithm~\ref{alg:recursive-qnn}.

First of all, as each call of $\operatorname{HistogramEqualize}$ either produces two recursive
calls or terminates depending on $level$ variable, the call relation of any
invocation of  $\operatorname{HistogramEqualize}$ will form a balanced binary tree. For
clarity, we note as $\M^l_k$, $\M^g_k$ the corresponding $\M^l$, $\M^g$ used
for a depth $k$ node of the binary tree.

By the assumption of $\M^l$, $\M^g$, we have
$\frac1{\gamma}\le \frac{\sum\M^l_k}{\sum\M^g_k} \le \gamma$. At the leaf
nodes, the application will be at most $K$ number of times, hence the number of
entries in leaf nodes will be different by at most $\gamma^{2k}$ number of times.

What remains to be proved is that no two leaf nodes produce the same quantized
value. We create an auxiliary variable $D_n^k\in\{0, 1\}$ to record whether
a depth $k$ node is on the right branch of their depth $k-1$ parent.
We can prove that node $n$ will map values to the interval $\sum_{k} D_n^k
2^{k-1}$, by observing that at Line~\ref{line:recursive-qnn-return} of
Algorithm~\ref{alg:recursive-qnn}, $\frac12$ will only be added if the right branch
of the call tree is taken.

As $\sum_{k} D_n^k 2^{k-1}$ is unique for all nodes, we complete the proof.
$\hfill\square$

\section{Quantization of GRU}
We first investigate the quantization of GRU as it is structurally simpler.
The basic structure of GRU cell may be described as follows:
\begin{align*}
\z_t &= \sigma (\W_z \cdot [\h_{t-1}, \x_t]) \\
\bs{r}_t &= \sigma (\W_r \cdot [\h_{t-1}, \x_t]) \\
\widetilde{\h_t} &= \tanh (\W \cdot [\bs{r}_t \circ \h_{t-1}, \x_t]) \\
\h_t &= (\bs{1} - \z_t) \circ \h_{t-1} + \z_t \circ \widetilde{\h_t}
\text{,}
\end{align*}
where $\bs{1}$ is a vector with all entries being 1, $\sigma$ stands for the sigmoid function, ``$\cdot$'' stands for the dot product, $[\x,\y]$ stands for the concatenation of two vectors $\x$ and $\y$, and $\circ$ stands for the Hadamard product.

Recall that to benefit from the speed advantage of bit convolution kernels, we
need to multiply the two matrix inputs in low bit forms, such that the dot product can be calculated
by bitwise operation.
For plain feed forward neural networks, as the convolutions take up most of
computation time, we can get decent acceleration by the quantization of inputs of
convolutions and their weights.
But when it comes to more complex structures like GRU, we need to check the
bitwidth of each interlink.

Except for matrix multiplications needed to compute $\z_t$, $\bs{r}_t$ and
$\widetilde{\h_t}$, the gate structure of $\widetilde{\h_t}$ and $\h_t$
brings in the need for element-wise multiplication. As the outputs of
the sigmoid function may have higher bitwidths, the element-wise multiplication
may need be done between floating point numbers (or in higher bitwidth format).
As $\widetilde{\h_t}$ and $\h_t$ are also the inputs to computations at the next timestamp, and noting that a
quantized value multiplied by a quantized value will have a larger bitwidth, we need to
insert additional quantization steps after element-wise multiplications.

Another problem with the quantization of GRU structure is that the value
ranges of gates are different. The range of $\tanh$ is $[-1, 1]$, which is different from the
value range $[0,1]$ of $\z_t$ and $\bs{r}_t$. If we want to preserve the original
activation functions, we will have the following quantization scheme:
\begin{align*}
\z_t &= \sigma (\W_z \cdot [\h_{t-1}, \x_t]) \\
\bs{r}_t &= \sigma (\W_r \cdot [\h_{t-1}, \x_t]) \\
\widetilde{\h_t} &= \tanh (\W \cdot [2Q_\mathrm{k}(\frac12(\bs{r}_t \circ
\h_{t-1})+\frac12)-1, \x_t])
\\
\h_t &=2Q_\mathrm{k}(\frac12((\bs{1} - \z_t) \circ \h_{t-1} + \z_t \circ
\widetilde{\h_t})+\frac12)-\bs{1}
\text{,}
\end{align*}
where we assume the weights $W_z, W_r, W$ have already been
quantized to the closed interval $[-1, 1]$, and input $\x_t$ have already been
quantized to $[-1, 1]$.

However, we note that the quantization function already has an affine
transform to shift the value range. To simplify the implementation, we replace
the activation functions of $\widetilde{\h_t}$ to be the sigmoid function, so
that $(\bs{1} - \z_t) \circ \h_{t-1} + \z_t \circ \widetilde{\h_t} \in [0,1]$.

Summarizing the above considerations, the quantized version of GRU could be
written as
\begin{align*}
\z_t &= \sigma (\W_z \cdot [\h_{t-1}, \x_t]) \\
\bs{r}_t &= \sigma (\W_r \cdot [\h_{t-1}, \x_t]) \\
\widetilde{\h_t} &= \sigma (\W \cdot [Q_\mathrm{k}(\bs{r}_t \circ \h_{t-1}), \x_t]) \\
h_t &=Q_\mathrm{k}((\bs{1} - \z_t) \circ \h_{t-1} + \z_t \circ \widetilde{\h_t})
\text{,}
\end{align*}
where we assume the weights $\W_z, \W_r, \W$ have already been
quantized to $[-1, 1]$, and input $\x_t$ have already been
quantized to $[0, 1]$.

\section{Quantization of LSTM}
The structure of LSTM can be described as follows:
\begin{align*}
\f_t &= \sigma (\W_\mathrm{f} \cdot [\h_{t-1}, \x_t] + \bs{b}_{\mathrm{f}}) \\
\bs{i}_t &= \sigma (\W_\mathrm{i} \cdot [\h_{t-1}, \x_t] + \bs{b}_{\mathrm{i}}) \\
\widetilde{\C_t} &= \tanh (\W_\mathrm{C} \cdot [\h_{t-1}, \x_t] + \bs{b}_{\mathrm{i}}) \\
\C_t &= \f_t \circ \C_{t-1} + \bs{i}_t \circ \widetilde{\C_t} \\
\bs{o}_t &= \sigma (\W_\mathrm{o} \cdot [\h_{t-1}, \x_t] + \bs{b}_{\mathrm{o}}) \\
\h_t &= \bs{o}_t \circ \tanh (\C_t)
\end{align*}

Different from GRU, $C_t$ cannot be easily quantized, since the value has not been bounded by activation functions like $\tanh$. This difficulty
comes from structure design and cannot be alleviated without introducing extra facility to clip value
ranges.
But it can be noted that the computations involving $C_t$ are all
element-wise multiplications and additions, which may take much less time than
computing matrix products.
For this reason, we leave $C_t$ to be floating point numbers.

To simplify implementation, $\tanh$ activation for output may be changed to
the sigmoid function.

Summarizing above changes, the formula for quantized LSTM can be:
\begin{align*}
\f_t &= \sigma (\W_\mathrm{f} \cdot [\h_{t-1}, \x_t] + \bs{b}_{\mathrm{f}}) \\
\bs{i}_t &= \sigma (\W_\mathrm{i} \cdot [\h_{t-1}, \x_t] + \bs{b}_{\mathrm{i}}) \\
\widetilde{\C_t} &= \tanh (\W_\mathrm{C} \cdot [\h_{t-1}, \x_t] + \bs{b}_{\mathrm{i}}) \\
\C_t &= \f_t \circ \C_{t-1} + \bs{i}_t \circ \widetilde{\C_t} \\
\bs{o}_t &= \sigma (\W_\mathrm{o} \cdot [\h_{t-1}, \x_t] + \bs{b}_\mathrm{o}) \\
\h_t &= Q_\mathrm{k}(\bs{o}_t \circ \sigma (\C_t))
\text{,}
\end{align*}
where we assume the weights $\W_\mathrm{f}, \W_\mathrm{i}, \W_\mathrm{C}, \W_\mathrm{o}$ have already been
quantized to $[-1, 1]$, and input $\x_t$ have already been
quantized to $[0, 1]$.


\begin{thebibliography}{10}

\bibitem{krizhevsky2012imagenet}
Krizhevsky A, Sutskever I, Hinton G~E.
\newblock Imagenet classification with deep convolutional neural networks.
\newblock In {\em Proc. Advances in neural information processing systems}, Dec.
  2012, pp. 1097--1105.

\bibitem{zeiler2014visualizing}
Zeiler M~D, Fergus R.
\newblock Visualizing and understanding convolutional networks.
\newblock In {\em Proc. European Conference on Computer Vision}, Sep. 2014, pp.
  818--833.

\bibitem{girshick2014rich}
Girshick R, Donahue J, Darrell T, Malik J.
\newblock Rich feature hierarchies for accurate object detection and semantic
  segmentation.
\newblock In {\em Proc. IEEE conference on Computer Vision and
  Pattern Recognition}, Jun. 2014, pp. 580--587.

\bibitem{long2015fully}
Long J, Shelhamer E, Darrell T.
\newblock Fully convolutional networks for semantic segmentation.
\newblock In {\em Proc. IEEE Conference on Computer Vision and
  Pattern Recognition}, Jun. 2015, pp. 3431--3440.

\bibitem{hinton2012deep}
Hinton G, Deng L, Yu D, Dahl G~E, Mohamed A~r, Jaitly N, Senior A, Vanhoucke V,
  Nguyen P, Sainath T~N et~al.
\newblock Deep neural networks for acoustic modeling in speech recognition: The
  shared views of four research groups.
\newblock {\em Signal Processing Magazine, IEEE}, 2012, 29(6):82--97.

\bibitem{DBLP:conf/icassp/GravesMH13}
Graves A, Mohamed A, Hinton G~E.
\newblock Speech recognition with deep recurrent neural networks.
\newblock In {\em Proc. {IEEE} International Conference on Acoustics, Speech and
  Signal Processing (ICASSP)},
  May 2013, pp. 6645--6649.

\bibitem{mikolov2013distributed}
Mikolov T, Sutskever I, Chen K, Corrado G~S, Dean J.
\newblock Distributed representations of words and phrases and their
  compositionality.
\newblock In {\em Proc. Advances in neural information processing systems}, Dec.
  2013, pp. 3111--3119.

\bibitem{sutskever2014sequence}
Sutskever I, Vinyals O, Le Q~V.
\newblock Sequence to sequence learning with neural networks.
\newblock In {\em Proc. Advances in neural information processing systems}, Dec.
  2014, pp. 3104--3112.

\bibitem{bahdanau2014neural}
Bahdanau D, Cho K, Bengio Y.
\newblock Neural machine translation by jointly learning to align and
  translate.
\newblock {\em arXiv preprint arXiv:1409.0473}, 2014.

\bibitem{mnih2015human}
Mnih V, Kavukcuoglu K, Silver D, Rusu A~A, Veness J, Bellemare M~G, Graves A,
  Riedmiller M, Fidjeland A~K, Ostrovski G et~al.
\newblock Human-level control through deep reinforcement learning.
\newblock {\em Nature}, 2015, 518(7540):529--533.

\bibitem{silver2016mastering}
Silver D, Huang A, Maddison C~J, Guez A, Sifre L, Van Den~Driessche G,
  Schrittwieser J, Antonoglou I, Panneershelvam V, Lanctot M et~al.
\newblock Mastering the game of go with deep neural networks and tree search.
\newblock {\em Nature}, 2016, 529(7587):484--489.

\bibitem{he2016identity}
He K, Zhang X, Ren S, Sun J.
\newblock Identity mappings in deep residual networks.
\newblock In {\em Proc. the 14th European Conference Computer Vision (ECCV)},
  Oct. 2016, pp. 630--645.

\bibitem{Simonyan14c}
Simonyan K, Zisserman A.
\newblock Very deep convolutional networks for large-scale image recognition.
\newblock {\em CoRR}, 2014, abs/1409.1556.

\bibitem{szegedy2014going}
Szegedy C, Liu W, Jia Y, Sermanet P, Reed S~E, Anguelov D, Erhan D, Vanhoucke
  V, Rabinovich A.
\newblock Going deeper with convolutions.
\newblock In {\em Proc. {IEEE} Conference on Computer Vision and Pattern Recognition}, Jun. 2015, pp. 1--9.

\bibitem{he2015deep}
He K, Zhang X, Ren S, Sun J.
\newblock Deep residual learning for image recognition.
\newblock In {\em Proc. {IEEE} Conference on Computer Vision and Pattern
  Recognition}, Jun. 2016,
  pp. 770--778.

\bibitem{DBLP:journals/tc/GalalH11}
Galal S, Horowitz M.
\newblock Energy-efficient floating-point unit design.
\newblock {\em {IEEE} Trans. Computers}, 2011, 60(7):913--922.

\bibitem{hochreiter1997long}
Hochreiter S, Schmidhuber J.
\newblock Long short-term memory.
\newblock {\em Neural computation}, 1997, 9(8):1735--1780.

\bibitem{chung2014empirical}
Chung J, G{\"{u}}l{\c{c}}ehre {\c{C}}, Cho K, Bengio Y.
\newblock Empirical evaluation of gated recurrent neural networks on sequence
  modeling.
\newblock {\em CoRR}, 2014, abs/1412.3555.

\bibitem{pham2012neuflow}
Pham P~H, Jelaca D, Farabet C, Martini B, LeCun Y, Culurciello E.
\newblock Neuflow: Dataflow vision processing system-on-a-chip.
\newblock In {\em Proc. IEEE the 55th International
  Midwest Symposium on Circuits and Systems (MWSCAS)}, 2012, pp. 1044--1047.

\bibitem{chen2014diannao}
Chen T, Du Z, Sun N, Wang J, Wu C, Chen Y, Temam O.
\newblock Diannao: a small-footprint high-throughput accelerator for ubiquitous
  machine-learning.
\newblock In {\em Proc. Architectural Support for Programming Languages and Operating
  Systems (ASPLOS)}, Mar. 2014,
  pp. 269--284.

\bibitem{DBLP:journals/tc/LuoLLWZCXTC17}
Luo T, Liu S, Li L, Wang Y, Zhang S, Chen T, Xu Z, Temam O, Chen Y.
\newblock Dadiannao: {A} neural network supercomputer.
\newblock {\em {IEEE} Trans. Computers}, 2017, 66(1):73--88.

\bibitem{denton2014exploiting}
Denton E~L, Zaremba W, Bruna J, LeCun Y, Fergus R.
\newblock Exploiting linear structure within convolutional networks for
  efficient evaluation.
\newblock In {\em Proc. Advances in Neural Information Processing Systems}, Dec. 2014, pp.
  1269--1277.

\bibitem{jaderberg2014speeding}
Jaderberg M, Vedaldi A, Zisserman A.
\newblock Speeding up convolutional neural networks with low rank expansions.
\newblock In {\em Proc. British Machine Vision Conference (BMVC)}, Sep. 2014.

\bibitem{tai2015convolutional}
Tai C, Xiao T, Wang X, E W.
\newblock Convolutional neural networks with low-rank regularization.
\newblock {\em CoRR}, 2015, abs/1511.06067.

\bibitem{zhou2015exploiting}
Zhou S, Wu J, Wu Y, Zhou X.
\newblock Exploiting local structures with the kronecker layer in convolutional
  networks.
\newblock {\em CoRR}, 2015, abs/1512.09194.

\bibitem{novikov2015tensorizing}
Novikov A, Podoprikhin D, Osokin A, Vetrov D~P.
\newblock Tensorizing neural networks.
\newblock In {\em Proc. Advances in Neural Information Processing Systems}, Dec. 2015, pp. 442--450.

\bibitem{zhang2015accelerating}
Zhang X, Zou J, He K, Sun J.
\newblock Accelerating very deep convolutional networks for classification and
  detection.
\newblock {\em {IEEE} Transaction on Pattern Analysis and Machine Intelligence}, 2016,
  38(10):1943--1955.

\bibitem{anwar2015structured}
Anwar S, Hwang K, Sung W.
\newblock Structured pruning of deep convolutional neural networks.
\newblock {\em CoRR}, 2015, abs/1512.08571.

\bibitem{han2015learning}
Han S, Pool J, Tran J, Dally W~J.
\newblock Learning both weights and connections for efficient neural network.
\newblock In {\em Proc. Advances in Neural Information Processing Systems}, Dec. 2015, pp. 1135--1143.

\bibitem{han2015deep}
Han S, Mao H, Dally W~J.
\newblock Deep compression: Compressing deep neural network with pruning,
  trained quantization and huffman coding.
\newblock {\em CoRR}, 2015, abs/1510.00149.

\bibitem{liu2015sparse}
Liu B, Wang M, Foroosh H, Tappen M~F, Pensky M.
\newblock Sparse convolutional neural networks.
\newblock In {\em Proc. {IEEE} Conference on Computer Vision and Pattern Recognition (CVPR)}, Jun. 2015, pp. 806--814.

\bibitem{cheng2015exploration}
Cheng Y, Yu F~X, Feris R~S, Kumar S, Choudhary A~N, Chang S.
\newblock An exploration of parameter redundancy in deep networks with
  circulant projections.
\newblock In {\em Proc. {IEEE} International Conference on Computer Vision}, Dec. 2015, pp.
  2857--2865.

\bibitem{chen2015compressing}
Chen W, Wilson J~T, Tyree S, Weinberger K~Q, Chen Y.
\newblock Compressing neural networks with the hashing trick.
\newblock In {\em Proc. the 32nd International Conference on Machine
  Learning}, Jul. 2015, pp.
  2285--2294.

\bibitem{DBLP:conf/kdd/ChenWTWC16}
Chen W, Wilson J~T, Tyree S, Weinberger K~Q, Chen Y.
\newblock Compressing convolutional neural networks in the frequency domain.
\newblock In {\em Proc. the 22nd International
  Conference on Knowledge Discovery and Data Mining}, Aug. 2016, pp. 1475--1484.

\bibitem{anguita2011fpga}
Anguita D, Carlino L, Ghio A, Ridella S.
\newblock A fpga core generator for embedded classification systems.
\newblock {\em Journal of Circuits, Systems, and Computers}, 2011,
  20(02):263--282.

\bibitem{vanhoucke2011improving}
Vanhoucke V, Senior A, Mao M~Z.
\newblock Improving the speed of neural networks on cpus.
\newblock In {\em Proc. Deep Learning and Unsupervised Feature Learning Workshop,
  NIPS}, Dec. 2011.

\bibitem{alvarez2016efficient}
Alvarez R, Prabhavalkar R, Bakhtin A.
\newblock On the efficient representation and execution of deep acoustic
  models.
\newblock In {\em Proc. the 17th Annual Conference of the International
  Speech Communication Association}, Sep. 2016, pp. 2746--2750.

\bibitem{zen2016fast}
Zen H, Agiomyrgiannakis Y, Egberts N, Henderson F, Szczepaniak P.
\newblock Fast, compact, and high quality {LSTM-RNN} based statistical
  parametric speech synthesizers for mobile devices.
\newblock In {\em Proc. the 17th Annual Conference of the International
  Speech Communication Association, San Francisco}, Sep. 2016, pp. 2273--2277.

\bibitem{gong2014compressing}
Gong Y, Liu L, Yang M, Bourdev L~D.
\newblock Compressing deep convolutional networks using vector quantization.
\newblock {\em CoRR}, 2014, abs/1412.6115.

\bibitem{merolla2016deep}
Merolla P, Appuswamy R, Arthur J~V, Esser S~K, Modha D~S.
\newblock Deep neural networks are robust to weight binarization and other
  non-linear distortions.
\newblock {\em CoRR}, 2016, abs/1606.01981.

\bibitem{gupta2015deep}
Gupta S, Agrawal A, Gopalakrishnan K, Narayanan P.
\newblock Deep learning with limited numerical precision.
\newblock {\em arXiv preprint arXiv:1502.02551}, 2015.

\bibitem{DBLP:journals/corr/CourbariauxB16}
Courbariaux M, Bengio Y.
\newblock Binarynet: Training deep neural networks with weights and activations
  constrained to +1 or -1.
\newblock {\em CoRR}, 2016, abs/1602.02830.

\bibitem{wu2015quantized}
Wu J, Leng C, Wang Y, Hu Q, Cheng J.
\newblock Quantized convolutional neural networks for mobile devices.
\newblock {\em CoRR}, 2015, abs/1512.06473.

\bibitem{kim2016bitwise}
Kim M, Smaragdis P.
\newblock Bitwise neural networks.
\newblock {\em CoRR}, 2016, abs/1601.06071.

\bibitem{DBLP:conf/nips/HubaraCSEB16}
Hubara I, Courbariaux M, Soudry D, El{-}Yaniv R, Bengio Y.
\newblock Binarized neural networks.
\newblock In {\em Proc. Advances in Neural Information Processing Systems}, Dec. 2016, pp. 4107--4115.

\bibitem{rastegari2016xnor}
Rastegari M, Ordonez V, Redmon J, Farhadi A.
\newblock Xnor-net: Imagenet classification using binary convolutional neural
  networks.
\newblock In {\em Proc. the 14th European Conference Computer Vision},
  Oct. 2016, pp. 525--542.

\bibitem{hinton2012neural}
Hinton G, Srivastava N, Swersky K.
\newblock Neural networks for machine learning.
\newblock {\em Coursera, video lectures}, 2012, 264.

\bibitem{bengio2013estimating}
Bengio Y, L{\'{e}}onard N, Courville A~C.
\newblock Estimating or propagating gradients through stochastic neurons for
  conditional computation.
\newblock {\em CoRR}, 2013, abs/1308.3432.

\bibitem{hwang2014fixed}
Hwang K, Sung W.
\newblock Fixed-point feedforward deep neural network design using weights +1,
  0, and -1.
\newblock In {\em Proc. {IEEE} Workshop on Signal Processing Systems}, Oct. 2014, pp. 174--179.

\bibitem{shin2016fixed}
Shin S, Hwang K, Sung W.
\newblock Fixed-point performance analysis of recurrent neural networks.
\newblock In {\em Proc. {IEEE} International Conference on Acoustics, Speech and
  Signal Processing (ICASSP)}, Mar.
  2016, pp. 976--980.

\bibitem{hubara2016quantized}
Hubara I, Courbariaux M, Soudry D, El{-}Yaniv R, Bengio Y.
\newblock Quantized neural networks: Training neural networks with low
  precision weights and activations.
\newblock {\em CoRR}, 2016, abs/1609.07061.

\bibitem{miyashita2016convolutional}
Miyashita D, Lee E~H, Murmann B.
\newblock Convolutional neural networks using logarithmic data representation.
\newblock {\em arXiv preprint arXiv:1603.01025}, 2016.

\bibitem{zhou2016dorefa}
Zhou S, Wu Y, Ni Z, Zhou X, Wen H, Zou Y.
\newblock Dorefa-net: Training low bitwidth convolutional neural networks with
  low bitwidth gradients.
\newblock {\em CoRR}, 2016, abs/1606.06160.

\bibitem{abaditensorflow}
Abadi M, Agarwal A, Barham P, Brevdo E, Chen Z, Citro C, Corrado G~S, Davis A,
  Dean J, Devin M et~al.
\newblock Tensorflow: Large-scale machine learning on heterogeneous systems.
\newblock {\em Software available from tensorflow.org}, 2015.

\bibitem{andri2016yodann}
Andri R, Cavigelli L, Rossi D, Benini L.
\newblock Yodann: An ultra-low power convolutional neural network accelerator
  based on binary weights.
\newblock In {\em Proc. {IEEE} Computer Society Annual Symposium on VLSI}, Jul. 2016, pp. 236--241.

\bibitem{lee2016fpga}
Lee M, Hwang K, Park J, Choi S, Shin S, Sung W.
\newblock Fpga-based low-power speech recognition with recurrent neural
  networks.
\newblock In {\em Proc. {IEEE} International Workshop on Signal Processing
  Systems}, Oct. 2016, pp.
  230--235.

\bibitem{courbariaux2015binaryconnect}
Courbariaux M, Bengio Y, David J.
\newblock Binaryconnect: Training deep neural networks with binary weights
  during propagations.
\newblock In {\em Proc. Advances in Neural Information Processing Systems}, Dec. 2015, pp. 3123--3131.

\bibitem{DBLP:conf/icml/SaxeKCBSN11}
Saxe A~M, Koh P~W, Chen Z, Bhand M, Suresh B, Ng A~Y.
\newblock On random weights and unsupervised feature learning.
\newblock In {\em Proc. the 28th International Conference on Machine
  Learning},
  Jun. 2011, pp. 1089--1096.

\bibitem{giryes2015deep}
Giryes R, Sapiro G, Bronstein A~M.
\newblock Deep neural networks with random gaussian weights: A universal
  classification strategy?
\newblock {\em IEEE Transactions on Signal Processing}, 2015,
  64(13):3444--3457.

\bibitem{DBLP:conf/siggraph/Heckbert82}
Heckbert P~S.
\newblock Color image quantization for frame buffer display.
\newblock In {\em Proc. the 9th Annual Conference on Computer Graphics
  and Interactive Techniques}, Jul. 1982, pp. 297--307.

\bibitem{mallows1991another}
Mallows C.
\newblock Another comment on o'cinneide.
\newblock {\em The American Statistician}, 1991, 45(3):257.

\bibitem{ioffe2015batch}
Ioffe S, Szegedy C.
\newblock Batch normalization: Accelerating deep network training by reducing
  internal covariate shift.
\newblock {\em arXiv preprint arXiv:1502.03167}, 2015.

\bibitem{netzer2011reading}
Netzer Y, Wang T, Coates A, Bissacco A, Wu B, Ng A~Y.
\newblock Reading digits in natural images with unsupervised feature learning.
\newblock In {\em Proc. Workshop on deep learning and unsupervised feature
  learning, NIPS}, volume 2011, 2011.

\bibitem{russakovsky2015imagenet}
Russakovsky O, Deng J, Su H, Krause J, Satheesh S, Ma S, Huang Z, Karpathy A,
  Khosla A, Bernstein M et~al.
\newblock Imagenet large scale visual recognition challenge.
\newblock {\em International Journal of Computer Vision}, 2015,
  115(3):211--252.

\bibitem{gysel2016hardware}
Gysel P, Motamedi M, Ghiasi S.
\newblock Hardware-oriented approximation of convolutional neural networks.
\newblock {\em CoRR}, 2016, abs/1604.03168.

\bibitem{taylor2003penn}
Taylor A, Marcus M, Santorini B.
\newblock {\em The Penn Treebank: An Overview}, pp. 5--22.
\newblock Springer Netherlands, Dordrecht, 2003.

\end{thebibliography}
\end{document}